 \documentclass[final,5p,times,twocolumn]{elsarticle}

\usepackage{amssymb}
\usepackage{lipsum}
\usepackage{algorithmic}
\usepackage{graphicx}
\usepackage{textcomp}
\usepackage{hyperref}
\usepackage{bbold}
\usepackage{tabularray}
\usepackage[normalem]{ulem}

\usepackage{amsthm, amsmath}

\newcommand{\etal}{\textit{et al.}}

\journal{Biomedical Signal Processing and Control}

\begin{document}

\begin{frontmatter}



\title{An efficient framework based on large foundation model for cervical cytopathology whole slide image screening}


\author[first]{Jialong Huang}
\affiliation[first]{organization={School of Computer Science and Engineering, Central South University},
            city={Changsha},
            postcode={410083}, 
            country={China}}
\affiliation[second]{organization={School of Automation, Central South University},
            city={Changsha},
            postcode={410083}, 
            country={China}}
\author[first]{Gaojie Li}
\author[first]{Shichao Kan}
\ead{kanshichao@csu.edu.cn}
\author[second]{Jianfeng Liu}
\ead{ljf@csu.edu.cn}
\author[first]{Yixiong Liang\corref{corresponding}}
\ead{yxliang@csu.edu.cn}
\cortext[corresponding]{Corresponding author.}

\begin{abstract}
Current cervical cytopathology whole slide image (WSI) screening primarily relies on detection-based approaches, which are limited in performance due to the expense and time-consuming annotation process. Multiple Instance Learning (MIL), a weakly supervised approach that relies solely on bag-level labels, can effectively alleviate these challenges. Nonetheless, MIL commonly employs frozen pretrained models or self-supervised learning for feature extraction, which suffers from low efficacy or inefficiency. In this paper, we propose an efficient framework for cervical cytopathology WSI classification using only WSI-level labels through unsupervised and weakly supervised learning. Given the sparse and dispersed nature of abnormal cells within cytopathological WSIs, we propose a strategy that leverages the pretrained foundation model to filter the top$k$ high-risk patches. Subsequently, we suggest parameter-efficient fine-tuning (PEFT) of a large foundation model using contrastive learning on the filtered patches to enhance its representation ability for task-specific signals. By training only the added linear adapters, we enhance the learning of patch-level features with substantially reduced time and memory consumption. Experiments conducted on the CSD and FNAC 2019 datasets demonstrate that the proposed method enhances the performance of various MIL methods and achieves state-of-the-art (SOTA) performance. The code and trained models are publicly available at \url{https://github.com/CVIU-CSU/TCT-InfoNCE}.
\end{abstract}




\end{frontmatter}




\section{Introduction}
\label{sec:introduction}

 According to global cancer statistics, cervical cancer ranks as the fourth most prevalent cancer globally and the second most common among women, particularly in developing and low-income countries, where it claims approximately 311,000 lives annually~\cite{ferlay2021cancer}.
 Over the past 60 years, cervical cytology screening has significantly contributed to reducing mortality rates associated with cervical cancer~\cite{marchevsky1996image}. Leveraging the advancements in digital pathology, there has been extensive research into Computer-Aided Diagnosis (CAD) systems, which aid healthcare professionals in tracking cervical cancer. 

\begin{figure}
    \centering
    \includegraphics[width=9cm]{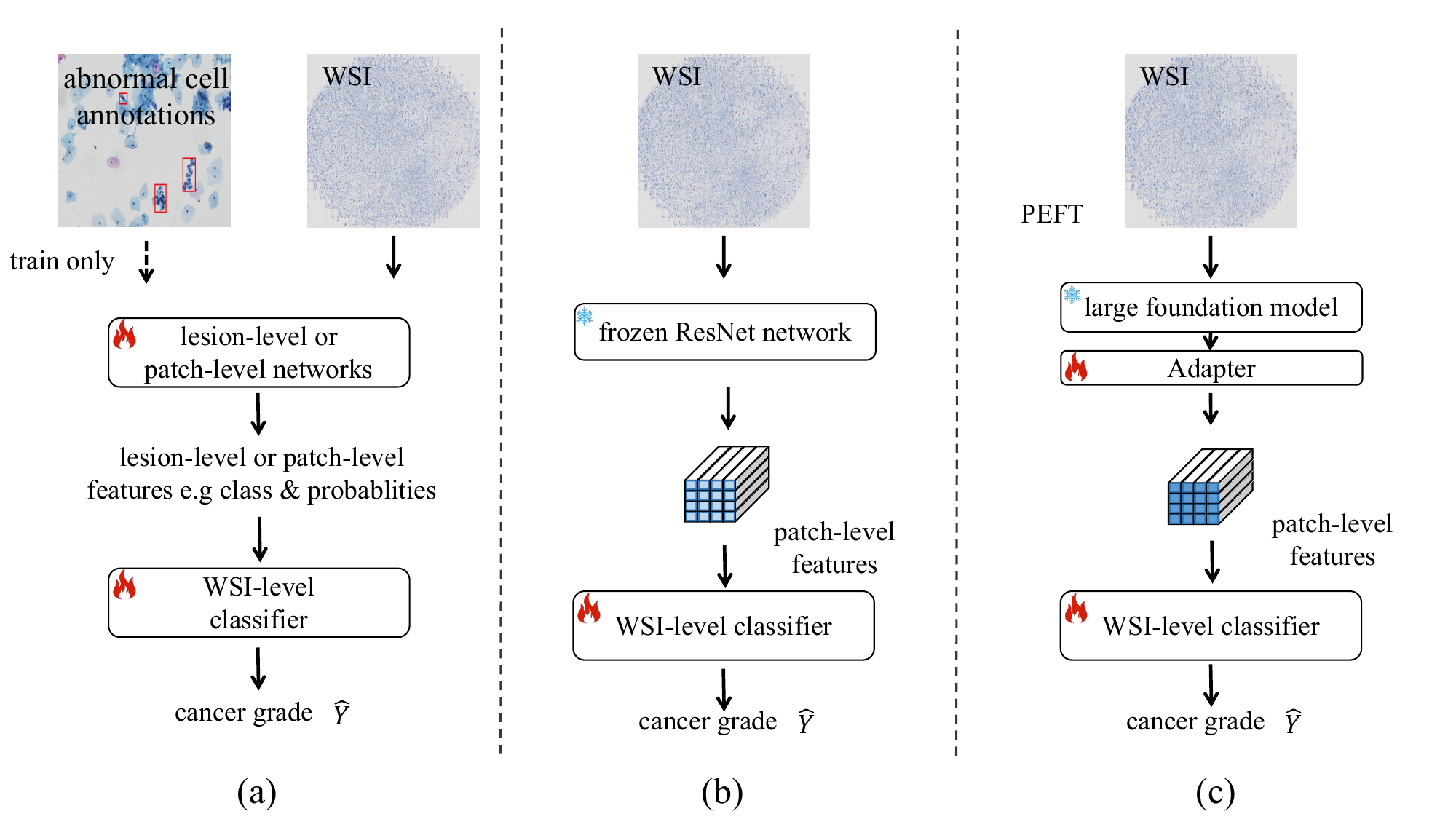}    \caption{Three deep learning paradigms for automated diagnosis on cervical cytopathology WSI: \textbf{(a)} Detection-based cancer screening framework based on abnormal cell annotations; \textbf{(b)} MIL methods for cancer screening with patch-level features extracted from frozen ResNet model; \textbf{(c)} Our proposed cancer screening with large foundation model adapted to cervical cytopathology WSI dataset.}
    \label{fig:intro}
\end{figure}

 Current research on cervical cytopathology whole slide image (WSI) screening primarily employs a detection-based framework, which consists of a detection model and a classification model~\cite{cao2021novel,lin2021dual,zhu2021hybrid,cao2023patch,wang2024artificial,cheng2021robust,geng2022learning,li2023novel}.
 Beyond WSI-level labels, these methods necessitate expert annotation at the lesion level to create a cervical cancer detection dataset. As shown in Fig.~\ref{fig:intro}(a), supervised training is then conducted on lesion-level or patch-level networks using this detection dataset. Subsequently, the fine-grained features extracted from these networks, such as detected abnormal cells, patch-level classifications, and associated probabilities, are fed into a WSI-level classifier for cancer screening. Some studies have achieved promising performance in cancer screening and demonstrated good interpretability for detecting certain lesions~\cite{lin2021dual,geng2022learning}. 
 However, annotating lesion-level datasets is a time-consuming and labor-intensive task for cytologists, and it carries the risk of missed or incorrect labels.
 
 The establishment of a detection-free framework, reliant solely on WSI-level labels that can be directly derived from clinical diagnoses, significantly enhances the model’s generalizability and practical utility. In the realm of histopathology WSI research, the classification of WSIs is often treated as a weakly supervised problem and the multiple instance learning (MIL) method has emerged as a paradigm for histopathology WSI screening, comprising a patch-level feature extractor and a WSI-level feature aggregator~\cite{ilse2018attention,lu2021data,shao2021transmil,tang2023multiple}.  Traditionally, frozen pre-trained models such as ResNet~\cite{he2016deep} are used to extract features, as illustrated in Fig.~\ref{fig:intro}(b); however, the domain gap between natural images and pathology images can lead to suboptimal results. Some works have proposed training image encoders in an unsupervised manner to enhance the learning of patch features~\cite{cao2023detection}; however, they overlook the computational overhead introduced by the gigapixel size of WSIs. 
 
 In this paper, we propose an efficient-parameter fine-tuning (PEFT) approach through contrastive learning to develop a cervical-specific feature extractor, shown in Fig.~\ref{fig:intro}(c). In the background of the rapid advancement of large foundation models, we suppose that leveraging the generalization and representational capabilities of these models can significantly enhance the representation of patches. In light of the domain gap between the pretraining datasets of these models and cervical images, as well as the considerable computational burden associated with fine-tuning large foundation models, PEFT can preserve the generalization of foundation models while enhancing their representational capabilities for cervical images. Additionally, the utilization of all patches derived from gigapixel WSIs for contrastive training is still impractical due to the considerable time costs. To address this, we propose a mean pooling (MP)-based approach for the preliminary segregation of high-risk patches, achieving a balance between efficacy and efficiency. Moreover, given the nature of cytopathology image screening, characterized by its scattered lesions and a paucity of structured information, focusing the feature extractor on high-risk patches is likely to enhance performance by emphasizing lesion-related features.
 
 Our contributions can be summarized in three main points:
 \begin{enumerate}
    \item We propose to adapt large foundation models by PEFT to learn a domain-specific feature extractor for cervical cytopathology WSI.
    \item We introduce an MP-based approach to filter out high-risk patches in the slides, addressing the challenges of unsupervised training caused by WSI super-resolution.
    \item We have demonstrated that the frozen foundation model with adapter effectively enhances the performance of various MIL methods through experiments conducted on a large cervical WSI dataset. Our method consistently achieves superior overall performance compared to state-of-the-art (SOTA) detection-based and detection-free methods.
\end{enumerate}

\section{Related work}
\label{sec:related}

\subsection{Cervical WSI Classification}
    Timely and accurate diagnosis through early cervical cytology screening enhances patients' survival rates. To improve diagnosis efficiency, the automation of cervical cytopathology WSI screening has been widely investigated. Deep learning-based cervical cancer screening emerging as a promising research direction with high performance~\cite{jiang2023systematic}. 
    Due to the gigapixel resolutions of WSIs, the WSI needs to be divided into equal-sized images, called patches, for processing by the neural network. Current approaches to cervical cytopathology WSI classification mainly fall into two categories: detection-based and detection-free methods.

    \textbf{Detection-based WSI-level classification.}
    The problem of cervical abnormal detection has been extensively studied thanks to modern object detection techniques~\cite{xiang_novel_2020,liang_comparison_2021,chen_task_2022, greenspan_robust_2023, chai_dpd-net_2024,song_cell_2024,liang2023exploring}. However, these methods focus primarily on detection performance and do not further investigate the classification performance of entire WSIs. With the increasing detection performance, several approaches have been proposed to use lesion features provided by the detection network to further train a cervical cytopathology WSI classifier, ultimately facilitating the comprehensive cancer screening of WSIs~\cite{cao2021novel,cheng2021robust,lin2021dual,zhu2021hybrid,geng2022learning}.
    
    Some works~\cite{cao2021novel,lin2021dual,zhu2021hybrid,cao2023patch, wang2024artificial} trained object detection model with lesion-level annotations firstly and then following the diagnosis procedures of cytologists, detected abnormal cells and aggregated corresponding lesion features for cancer screening. Instead, Cheng~\etal~\cite{cheng2021robust} employed two convolutional networks to extract patch-level features from high- and low-resolution WSIs for further aggregation. In particular, Geng~\etal~\citep{geng2022learning} developed a two-stage framework with lesion-level and patch-level loss in the first stage. However, directly using lesion-level features for classification may overlook the interconnections between lesion cells and the beneficial context information at the patch level for lesion detection~\cite{li2023novel}. Li~\etal~\cite{li2023novel} harnessed lesion-informative queries derived from the detection network Deformable DETR~\cite{zhu2020deformable}, aggregating these queries to generate patch-level features for further WSI-level classification.

    However, the above methods are based on abnormal cell labels which need cytologists to scan the gigapixel WSIs and annotate manually. Manual annotation is a tedious and labor-intensive task for cytologists, making it difficult to construct large-scale WSI datasets. What's worse, the performance of detection-based methods is largely dependent on the detection of abnormal cells and overlooks the overall information provided by the whole slide, leading to suboptimal classification performance.  Therefore, research on detection-free methods is necessary to overcome these challenges.
    
    \textbf{Detection-free WSI-level classification.} 
    The efficacy of detection-free methods for WSI classification depends on the extracted patch features’s capacity to encapsulate both lesion-specific and contextual information within the patch. Specifically, LESS~\cite{zhao2024less} employed Variational Positive-Unlabeled (VPU) learning~\cite{chen2020variational} by training a VPU encoder to directly extract patch features. Additionally, Cao~\etal ~\cite{cao2023detection} employed contrastive learning to train a coarse and a fine encoder for the extraction of patch features. Our work aims to enhance the representational ability of the patch-level feature extractor in an efficient manner, thereby addressing the computational overhead associated with gigapixel images and facilitating WSI-level classification.
    
\subsection{MIL in WSI Classification} 
    Similar to cytopathology WSI, histopathology WSI automatic diagnosis also faces difficulties with large pixel size and limited annotation. MIL methods were developed to address these challenges, and have achieved good performance in analyzing histopathology
    WSI of various diseases with only WSI-level labels~\cite{ilse2018attention,li2021dual,lu2021data,shao2021transmil,tang2023multiple,tang2024feature,zhang2022dtfd}.
    In the framework of MIL, a WSI is partitioned into patches which are then transformed into low-dimensional patch features by a frozen image encoder. 
    
    Generally, MIL can be divided into two categories: instance-based and embedding-based methods~\cite{li2021dual}. Instance-based methods score the patches for risk assessment and then integrate the patch scores by pooling operation. Characterized by the scarcity of trainable parameters and simplistic architectural designs, these methods exhibit reduced computational costs but compromised performance. In contrast, embedding-based methods first interact with patch-level features in different ways and then further integrate the combined features for cancer classification. This category of methods has seen extensive development, with approaches like AB-MIL~\cite{ilse2018attention} using gated attention, CLAM~\cite{lu2021data} further incorporates instance-level clustering and TransMIL~\cite{shao2021transmil} employing self-attention and PPEG modules. Additionally, MHIM-MIL~\cite{tang2023multiple} enhanced performance by incorporating hard sampling and siamese network strategies on top of the previous methods.

    The success of the mainstream MIL methods in histopathology WSIs screening is greatly attributed to the image feature extraction capabilities of the ResNet~\cite{he2016deep} model pretrained on ImageNet~\cite{russakovsky2015imagenet}. However, there is a gap between the images of ImageNet and pathology patches which limits the performance of MIL methods. To span this gap, some work~\cite{lin2021dual,cao2023detection} used self-supervised contrastive learning to enhance the ResNet network’s ability to represent visual features of pathological slides. Cao~\etal~\cite{cao2023detection} followed MoCo v2~\cite{he2020momentum} to train ResNet50~\cite{he2016deep} with all patches from gigapixel WSIs, which is inefficient due to the large areas with minimal variation in the background and it entails an excessive consumption of memory and time resources. In contrast, Li~\etal~\cite{li2021dual} adopted random sampling to alleviate the issue caused by the large number of patches. Images of these two methods provided for contrastive learning contain too much background information or too little lesion information, which can negatively impact the effectiveness.

\subsection{Foundation Models}
    Recently, foundation models which are large models pretrained in large-scale datasets have emerged as a trend in many research fields. Due to general abilities and independence on extensive labeling, healthcare foundation models (HFMs) have also gained great interest and obtained impressive success~\cite{he2024foundation}. Specially, Based on LLaMA~\cite{touvron2023llama}, PMC-LLaMA~\cite{wu2024pmc} built up a large language model (LLM) for medicine beating LLaMA-2~\cite{touvron2023llama2} with fewer parameters various public medical question-answering benchmarks. Since healthcare data is inherently multimodal, therefore it is promising to construct a multimodal foundation model (MFM) to integrate modalities. Zhang~\etal~\cite{zhang2023biomed} introduced PMC-15M, a large-scale dataset encompassing kinds of biomedical image-text pairs, such as pathological images. Furthermore, a multimodal biomedical foundation model, BiomedCLIP~\cite{zhang2023biomed}, was pretrained on PMC-15M, showcasing exceptional performance across various tasks through zero-shot learning. In addition to foundation models encompassing a variety of tasks, domain-specific foundation models have also made significant strides. 
    
    In the realm of pathological image screening, there have emerged numerous large-scale foundation models~\cite{Lu_2023_CVPR, huang2023visual, radford2021learning, xu2024whole}. Especially, PLIP~\cite{huang2023visual} fine-tuned the pretrained contrastive language-image pretraining~\cite{radford2021learning} model on OpenPath which is a large dataset of 208,414 pathology image-text pairs collected from Twitter community. Moreover, Prov-GigaPath~\cite{xu2024whole}, a whole-slide pathology foundation model pretrained on 1.3 billion $256\times 256$ pathology image tiles from 171,189 whole slides, has achieved SOTA performance on several benchmarks. In response to the challenges posed by the large number of parameters of foundation models, we leverage the PEFT algorithm to adapt these models to cervical WSI and significantly enhance their representational capabilities in the context of cervical cytopathology WSI.

\section{Methods}\label{sec:methods}

\begin{figure*}
    \centering
    \includegraphics[width=16cm]{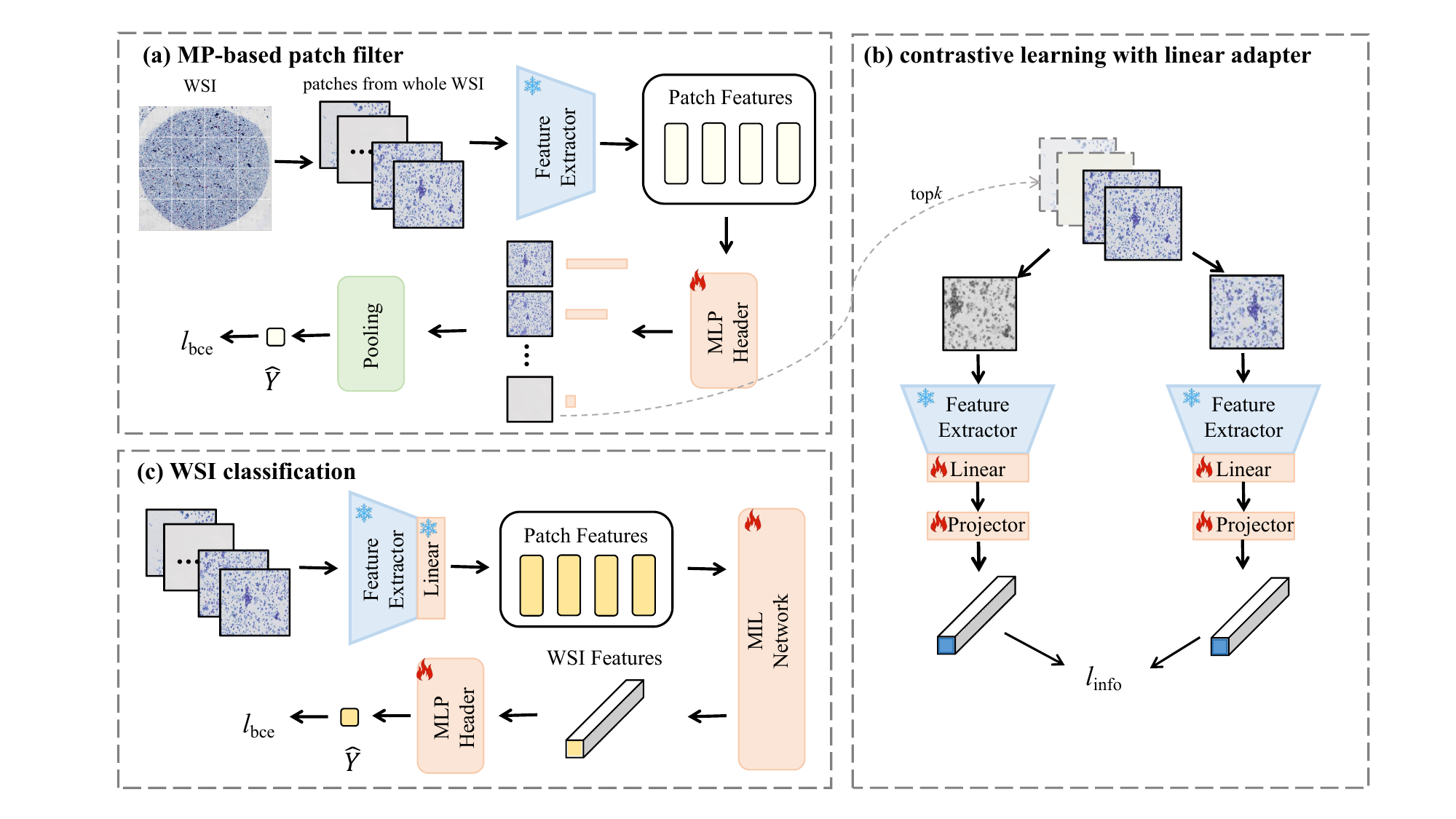}    \caption{Framework of our method. \textbf{(a)} top$k$ high-risk patches are selected by using filter strategy and used to train the frozen image encoder with adapter by contrastive learning. All patches are sent to the feature extractor with an adapter to extract patch features, and the patch features from one WSI are fed to the MIL network to make the final prediction. \textbf{(b)} Our filter strategy is training an MP-based method and using patch scores to filter patches. }
    \label{fig:framework}
\end{figure*}

\subsection{Background}
In scenarios of binary MIL classification, let $B=\{(p_1,y_1),...,(p_n,y_n)\} $ be a bag with instances $p_i \in X$ with labels $y_i \in \{0,1\}$. The bag label $Y$ is given by
\begin{equation}
    Y=c(X) = 
    \begin{cases} 
        0, & \text{iff} \sum{y_i}=0 \\
        1, & \text{otherwise}.
    \end{cases}
    \label{equation:mil1}
\end{equation}

In the context of MIL applied to WSIs, each WSI is conceptualized as a singular bag, wherein individual patches divided from WSI are regarded as instances. The labeling of the WSI is determined by the presence or absence of infection within these patches; if no infected patches are detected, the WSI is labeled as negative, whereas the presence of infection in any patch results in a positive label.
Normally, MIL uses an instance-level transformation $f$ and a bag-level transformation $g$ to predict the label of the bag:

\begin{equation}
    c(X) = g(f(p_1),...,f(p_n)).
    \label{equatiom:mil2}
\end{equation}

Based on the choice of $f$ and $g$, MIL can be divided into two categories~\cite{li2021dual}: 1) Instance-based approach. $f$ is an instance-level classifier to predict instance scores and $g$ is a pooling operator, such as max pooling and mean pooling. 2) Embedding-based approach. $f$ is a feature extractor that transforms instances into embedding and $g$ is a bag-level aggregator that outputs the bag score based on instances embedding. 
Compared to the instance-based method, The embedding-based method yields better accuracy, but it is harder to identify the pivotal instances that trigger the classification~\cite{li2021dual}. 

\subsection{Method Framework}
\label{ssec:m_o}
As illustrated in fig.\ref{fig:framework}, in the inference stage, consistent with the MIL approach, the process of our framework mainly comprises two components: the patch-level extractor and the WSI-level aggregator. In light of the disparity between natural images and cytopathology WSIs, we deviate from the conventional approach in histopathology WSI research~\cite{lu2021data}, which uses ResNet models pretrained on ImageNet in the MIL method. Instead, we train a feature extractor specifically for cervical cytopathology WSIs based on the image encoder from a large foundation model. By employing a combination of unsupervised and weakly supervised training methods for the patch-level extractor, we enhance the foundation model’s ability to represent images in the task domain under detection-free conditions. 

Training the patch-level extractor for cervical cytopathology WSI is a two-stage process, as illustrated within the dashed box in Fig.~\ref{fig:framework}. Firstly, we have patches divided from WSI, denoted as $X = \{p_1, p_2,..., p_n\}$, and a frozen image encoder $\phi_{b}$ pretrained on the large medical dataset. However, considering the presence of numerous similar and repetitive regions within WSI samples, especially in negative samples, and the unacceptable time and memory costs associated with unsupervised training on all patches~\cite{li2021dual}, we first employ weakly supervision learning to filter out top$k$ high-risk patches before tuning the image encoder, denoted as $\{p_{r1}, p_{r2},..., p_{rk}\}$. Secondly, to simplify the training process, we just regard high-risk patches from $N$ WSIs as individual images to construct an image dataset $\{x_1, x_2, ...x_{N\times k} \}$ and follow the practice of SimCLR~\cite{chen2020simple} to train the image encoder. 

Once the PEFT for the image encoder is complete, the entire network comprising the original network and the adapter module, is frozen as an offline module and employed to extract patch-level features, as shown in Fig.~\ref{fig:framework}(c). Then an embedding-based MIL approach trained solely on WSI-level labels is utilized for cancer screening and calculated the BCE loss. For the MIL training formulation of one WSI $X = \{p_1, p_2,..., p_n\}$ with label $Y$: 
\begin{align}
    z_i &= \phi_a( \phi_b(p_i)), \\
    \hat{Y} &= \phi_{mil}(z_1, ..., z_n),   \\
    \ell_{bce}(Y,\hat{Y}) &= Y\cdot \log{\hat{Y}} + (1-Y)\cdot \log(1-\hat{Y}).
\end{align}
where $z_i$ with shape $D\times 1$ is the visual representation of image $p_i$ whose shape is $H\times W \times 3$, $\hat{Y}$ is the prediction result, $\phi_{b}$ is the frozen image encoder, $\phi_{a}$ is the adapter module and $\phi_{mil}$ is the embedding-based MIL network.

\subsection{MP-based Patch Filter} 
\begin{figure}
    \centering
    \includegraphics[width=8.5cm]{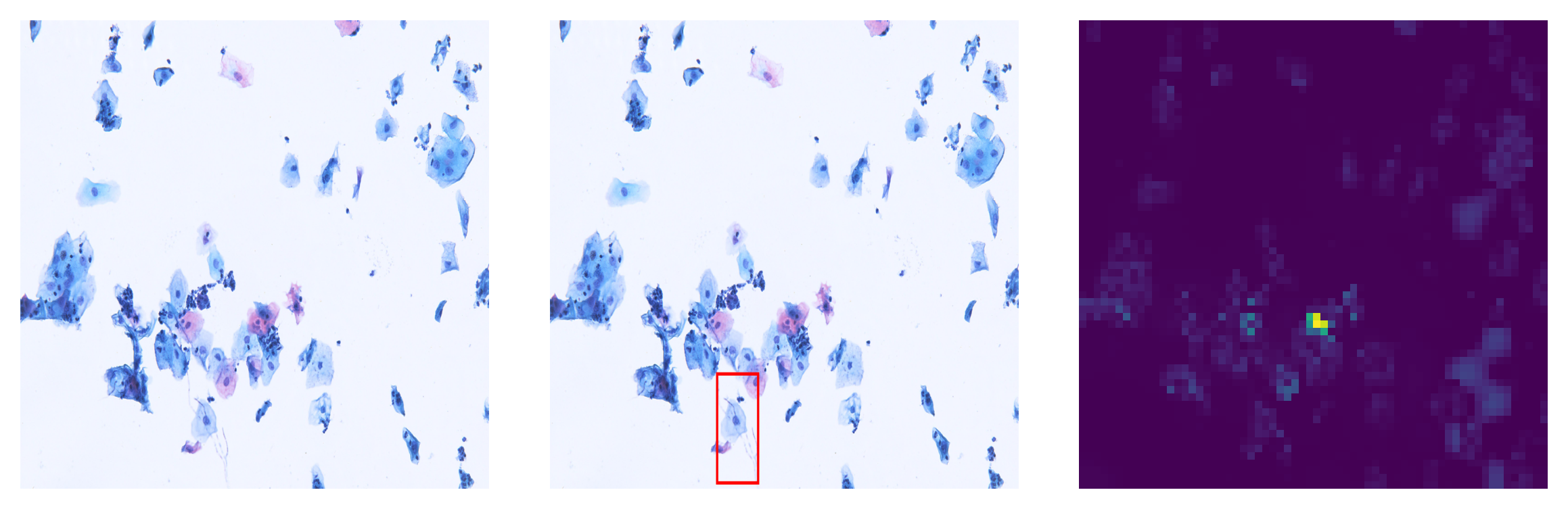}
    \includegraphics[width=8.5cm]{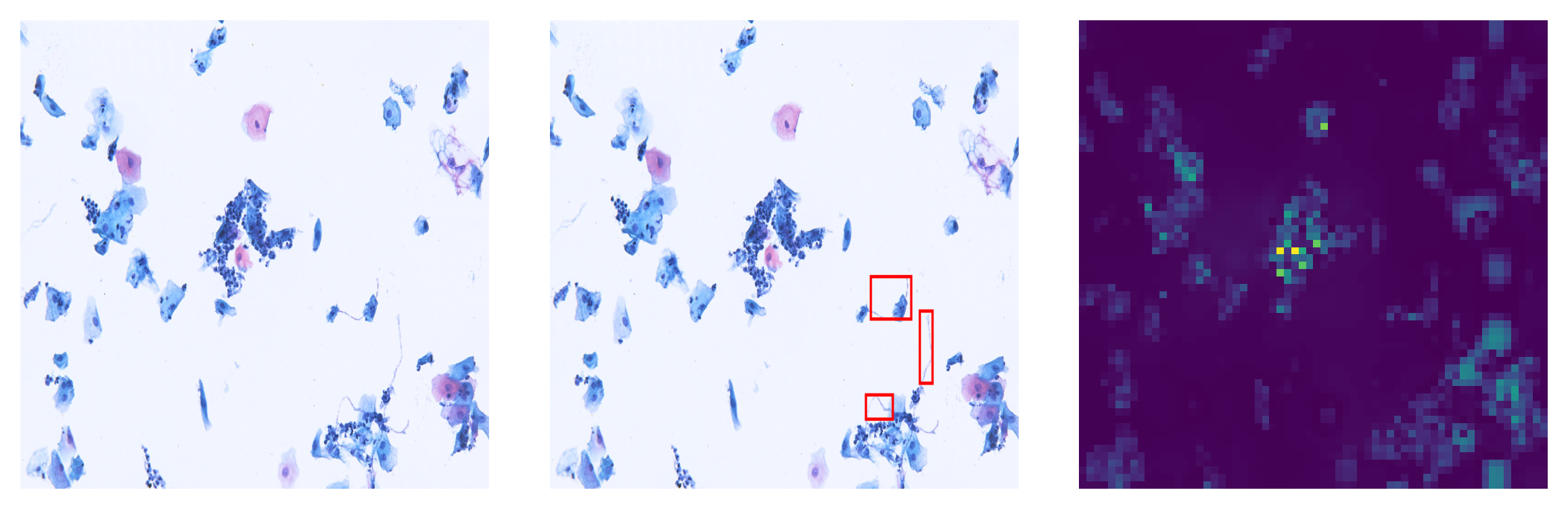}
    \caption{Visualization of the influence of image patches on the class token. The first column shows the original images, the second column highlights the lesion areas with boxes, and the third column displays the attention weights of the image patches concerning the class token. }
    \label{fig:bvis}
\end{figure}

Given the substantial size of cervical cytopathological images, which results in a large number of patches when split, the utilization of all patches for unsupervised training~\cite{cao2023detection} would entail an exorbitant time investment. Consequently, we propose a strategy to select high-risk patches for unsupervised training, thereby enhancing the image encoder’s proficiency in representing lesion areas. Ideally, the remaining patches should predominantly consist of lesion regions in positive samples and challenging-to-discriminate regions in negative samples. We observed that foundation models trained on large-scale datasets of image-text pairs exhibit robust image representation capabilities. From a visual analysis perspective, as illustrated in Fig.~\ref{fig:bvis}, even without specific training, the foundation model can effectively discern cellular structures. Quantitatively, employing features extracted from a frozen foundation model and applying straightforward yet interpretable MIL methods such as mean pooling and max pooling yields satisfactory performance. Consequently, we advocate for the adoption of an MP-based patch filter.

The process of filtering high-risk patches comprises a training phase and an inference phase of mean pooling MIL. In \eqref{equation:mil1} for mean pooling MIL,  $f$ represents the frozen foundation model equipped with a linear layer, while 
$g$ denotes the mean pooling operation. Given that the foundation model is fixed and no data augmentation is employed, we pre-extract the features of patches, significantly reducing the computational time. The objective of mean pooling MIL is to diagnose WSIs, and the network without the pooling operation is viewed as a patch-level classifier, as depicted in Fig.~\ref{fig:framework}(a). Subsequently, the top-K patches, which encompass lesion regions within positive WSIs and are difficult to differentiate in negative WSIs, are meticulously selected.

\subsection{Contrastive Learning with Linear Adaptation}
As mentioned before, the performance of the MIL method depends on the representational ability of the patch-level extractor. We propose contrastive learning with linear adaptation to efficiently train a domain-specific extractor, thereby enhancing the classification performance of the MIL method on cervical cytopathology WSIs. A linear adapter module, denoted as $\phi_a$, is appended to the end of the frozen image encoder, enhancing its task-specific proficiency while retaining the large model's generalization capacity.

In detail, one image $x_i$ in a batch undergoes two random data augmentation to produce two samples $\tilde{x}_{2i}, \tilde{x}_{2i+1}$. In contrast to SimCLR~\cite{chen2020simple}, random color jitter is excluded from our data augmentation protocol. The feature for $\tilde{x}_i$ is given by:
\begin{equation}
z^{'}_i = G(\phi_a(\phi_b(\tilde{x}_i))),
\end{equation}
where $z^{'}_i$ is vector with shape $D^{'}$, and $G$ is a projection head. 

Within a mini-batch of $M$ images, there are $2M$ samples after data augmentation. For each sample, the positive pair is the sample and its counterpart and the negative pair is the sample and the other $2(M-1)$ samples. Let ${\rm sim}(u,v) = u^{\rm{T}}v/||u||\cdot ||v|| $ and the contrast learning loss is
\begin{equation}
    \ell_{info} = - \frac{1}{2M} \sum_{j=1}^{2M} \log \frac{\exp ({{\rm sim}(z_j^{'}, z^{'}_{a(j)})} / \tau) }
        {\sum_{m=1}^{2M}{\mathbb{1}_{[m\neq j]} \exp ({\rm sim}(z^{'}_j, z^{'}_m)/ \tau) }}, \\
\end{equation}
where $\mathbb{1}_{[m\neq j]} \in [0,1]$ is an indicator function evaluating to 1 iff $m\neq j$, $\tau$ denotes a temperature parameter and $a(j)$ denotes the index associated with the same image after undergoing a different form of data augmentation. Specially, the frozen image extractor can be any CNN or transformer structures from the foundation model. 

\section{Experiment}
\label{sec:experiment}

\subsection{Dataset and experiments settings}

\textbf{Datasets.} We conduct extensive experiments on the Cervical Smear Dataset (CSD)~\cite{geng2022learning,liang2023exploring} and FNAC 2019~\cite{saikia2019comparative}. 
CSD~\cite{geng2022learning,liang2023exploring} contains 2625 cervical WSIs and the diagnosis given by pathologists. In total, there are 1542 negative WSIs and 1083 positive WSIs which include the sub-categories of cervical cancer: NILM, ASC-US, LSIL, ASC-H, HSIL, and SCC. For every WSI, the range of patches varies from 126 to 1395, each composed of $4,096\times 2,816$ pixels. The average number of these patches per WSI is calculated to be 758. Following the protocol of~\cite{geng2022learning}, 70\% of the data is used for training, and the remaining 30\% is used for testing. 

\begin{table}
\centering
\caption{The detailed class distribution of datasets.}
\label{table:dataset}
\scalebox{0.9}{
\begin{tblr}
{
  cells = {c},
  cell{2}{1} = {r=2}{},
  cell{4}{1} = {r=2}{},
  vline{3} = {-}{},
  hline{1-2,4,6} = {-}{},
}
Dataset   &       & Benign & Malignant & Total \\
CSD~\cite{geng2022learning, liang2023exploring}       & Train & 1080   & 759       & 1839  \\
          & Test  & 462    & 324       & 786   \\
FNAC 2019~\cite{saikia2019comparative} & Train & 81     & 88        & 169   \\
          & Test  & 18     & 25        & 43    
\end{tblr}
}

\end{table}

FNAC 2019~\cite{saikia2019comparative} comprises 212 fine-needle aspiration breast cell inspection images, which are categorized as benign (99) or malignant (113). We extend our method to the FNAC 2019 dataset, which encompasses high-resolution breast cytological images of $2,048 \times 1,536$. The dataset is randomly partitioned into an 80\% training subset and a 20\% testing subset. The detailed class distribution of datasets is shown in Table~\ref{table:dataset}.

\textbf{Evaluation metrics.} For the 2-category classification task on CSD, NILM subtype is the benign type and the remaining subtypes are the malignant type. To conduct quantitative evaluation, we adopt sensitivity (sens), specificity (spec), precision (prec), F1-scores and AUC as evaluation metrics. 

\textbf{Implementation details.} Before training, we resize all the patches from $4,096 \times 2,816$ on CSD and $256 \times 256$ on FNAC to $224 \times 224$ to reduce time and space costs. The framework of our method can be viewed as three stages: training the mean pooling MIL, image encoder, and embedding-based MIL methods. In the first stage, we use frozen ViT~\cite{dosovitskiy2020image} from the foundation model to extract the visual features of patches and train a patch-level classifier. After filtering top$k$ ($k=50$) high-risk patches, 80\% of the training WSI are used to train patch-level feature extractor by our proposed PEFT method. Lastly, the trained extractor is viewed as an offline module to extract better features for MIL methods. Learning rate, batch size, momentum, weight decay, and temperature for contrast learning are set to 0.6, 1024, 0.9, 1.0e-6, and 0.5, and SGD is used as the optimizer. Learning rate, batch size, and weight decay for MIL methods are set to 2e-4, 1, and 1e-5, and Adam with cosine annealing is the optimizer.

\subsection{Comparisons with the SOTA methods}

\begin{table}
\centering
\caption{Comparative results on CSD test. The bold and underlined number indicates the optimal specificity or precision performance under a specific constraint.}
\label{table:csd}
\scalebox{0.9}{
\begin{tblr}{
  cells = {c},
  cell{2}{1} = {r=6}{},
  cell{8}{1} = {r=4}{},
  vline{3} = {-}{},
  hline{1-2,8,12} = {-}{},
}
                & Methods    & AUC   & Sens  & Spec  & Prec  \\
{Detection\\-based} & Lin-0.90~\cite{lin2021dual}    & -   & 92.28 & 30.74 & 48.30  \\
                & Lin-0.95~\cite{lin2021dual}  & -   & 95.68 & 27.71 & 48.14 \\
                & Gen-0.90~\cite{geng2022learning}  & 98.25 & 94.44 & 90.91 & 87.93 \\
                & Gen-0.95~\cite{geng2022learning}  & 98.25 & 97.84 & 85.50  & 82.55 \\
                & Liang-0.90~\cite{liang2023exploring} & 99.12   & 94.44 & \uline{97.60}  & \uline{96.50}  \\
                & Liang-0.95~\cite{liang2023exploring} & 99.12   & 97.84 & 86.80  & 83.90  \\
{Detection\\-free}  & Cao-0.90~\cite{cao2023detection}    & 96.69 & 94.44 & 88.75 & 85.47 \\
                & Cao-0.95~\cite{cao2023detection}   & 96.69 & 97.84 & 72.08 & 71.08 \\
                & ours-0.90  & \textbf{99.69} & 94.44 & \textbf{98.92}  & \textbf{98.39} \\
                & ours-0.95  & \textbf{99.69} & 97.84 & 95.67 & 94.07 
\end{tblr}}
\end{table}

\begin{table}
\centering
\caption{Comparative results with pretrained model. MHIM(AB) and MHIM(TR) correspond to MHIM-AB-MIL and MHIM-TransMIL, respectively.}
\label{table:backbone}
\scalebox{0.9}{
\begin{tblr}{
  cells = {c},
  cell{2}{1} = {r=4}{},
  cell{6}{1} = {r=4}{},
  cell{10}{1} = {r=4}{},
  vline{3} = {-}{},
  hline{1-2,6,10,14} = {-}{},
}
Feature extractor                & MIL      & AUC            & Sens           & Spec           \\
ResNet50                     & AB-MIL~\cite{ilse2018attention}   & 96.77          & 95.06          & 91.35          \\
                             & TransMIL~\cite{shao2021transmil} & 98.51          & 97.53          & 91.99          \\
                             & MHIM(AB)~\cite{tang2023multiple} & 97.91          & 96.30          & 92.85          \\
                             & MHIM(TR)~\cite{tang2023multiple}& 98.51          & 92.59          & 95.02          \\
BiomedCLIP                   & AB-MIL~\cite{ilse2018attention}  & 97.04          & 93.21          & 93.50          \\
                             & TransMIL~\cite{shao2021transmil}& 97.76          & 94.14          & 91.99          \\
                             & MHIM(AB)~\cite{tang2023multiple}& 98.28          & 95.68          & 93.94          \\
                             & MHIM(TR)~\cite{tang2023multiple}& 98.00          & 96.91          & 90.26          \\
{BiomedCLIP\\+Adapter(ours)} & AB-MIL~\cite{ilse2018attention}  & 98.84          & 95.37          & \textbf{96.32} \\
                             & TransMIL~\cite{shao2021transmil} & 99.14          & \textbf{98.46} & 94.59          \\
                             & MHIM(AB)~\cite{tang2023multiple}& 99.17          & 96.91          & 95.45          \\
                             & MHIM(TR)~\cite{tang2023multiple}& \textbf{99.18} & 97.22          & 95.45          
\end{tblr}
}
\end{table}

To substantiate the efficacy of our proposed framework in cervical WSI classification, we first compare our approach against SOTA methods on the CSD dataset and a detection-free method. Furthermore, to validate the enhancement of the patch-level extractor we have trained on MIL methods, we contrast our framework with mainstream MIL approaches on CSD~\cite{geng2022learning,liang2023exploring} and FNAC 2019~\cite{saikia2019comparative}, employing pretrained models such as ResNet50~\cite{he2016deep} and BiomedCLIP~\cite{huang2023visual}.

\textbf{Comparison with cervical task-specific methods.} Compare our approach with two categories of methods for cervical cytopathology WSI classification: detection-based~\cite{lin2021dual,geng2022learning,liang2023exploring} and detection-free~\cite{cao2023detection}. In line with~\cite{geng2022learning}, we establish 0.90 and 0.95 as the threshold constraints for the `minimum required’ sensitivity. Model performance is assessed by comparing specificity and precision under these constraints. The results of the detection-based methods are from~\cite{liang2023exploring}, and the results of~\cite{cao2023detection} were reproduced by us based on the original paper and repository guidance. Specifically, we used the authors' pretrained ResNet50~\cite{he2016deep} model for feature extraction and applied their proposed MIL method for aggregation. 

As reported in Table~\ref{table:csd}, in the detection-based methods,~\cite{liang2023exploring} demonstrated superior performance in the classification of WSIs, attributable to its novel modules, which enhanced the robustness of lesion detection. Under both sensitivity constraints, it surpassed the other two methods. In the detection-free approach, our method exhibited substantial advancements over~\cite{cao2023detection} under both constraints. Compared to~\cite{liang2023exploring}, our method achieved comparable performance at low sensitivity and demonstrated an 8.87\% and 10.17\% enhancement in specificity and precision, respectively, at high sensitivity (equivalent to 97.84\%). The possible reason that~\cite{liang2023exploring}'s image encoder was tailored for the detection of abnormal cells, resulting in excessive sensitivity and a high incidence of false positives, particularly at high sensitivity levels where distinguishing normal cells is challenging. In terms of training process and performance evaluation, our method, which does not require annotations of abnormal cells, exhibits substantial improvements over various methods at high sensitivity, rendering it highly beneficial for practical applications.

\begin{figure}
    \centering
        \includegraphics[width=7.5cm]{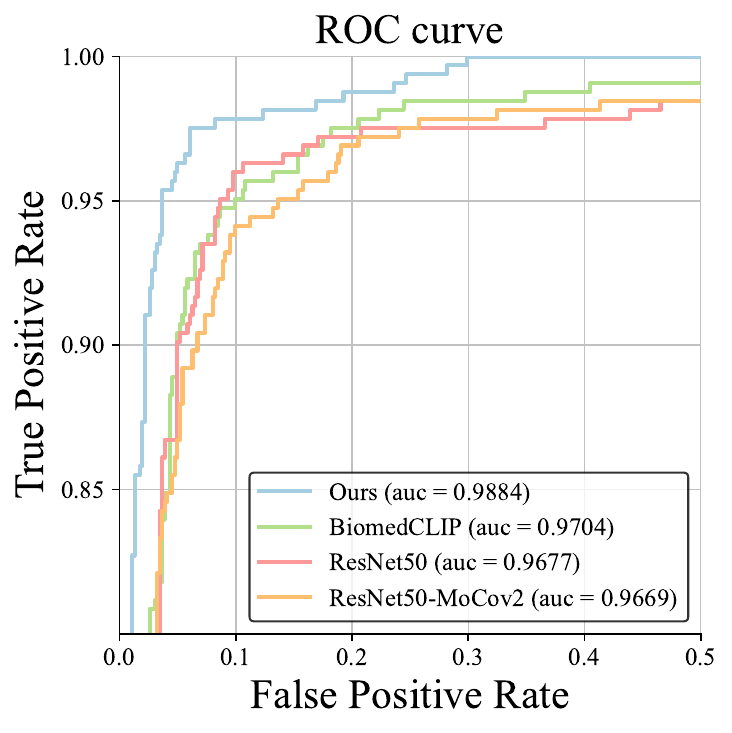}
    \caption{The ROC curves of different methods. The first three curves represent results obtained using BiomedCLIP-SimCLR, BiomedCLIP, and ResNet50 image encoders to extract features, which are then integrated through ABMIL to obtain the final results. The last curve follows the approach described in~\cite{cao2023detection} for WSI classification.}
    \label{fig:roc}
\end{figure}

\textbf{Comparison with MIL methods with frozen pretrained models on CSD.}
Essentially, our method focuses on training a cervical image-specific image encoder for the MIL method. To demonstrate the enhanced capabilities of our trained image encoder for representing cervical images, we conducted a comparative analysis across four MIL methods: AB-MIL~\cite{ilse2018attention}, TransMIL~\cite{shao2021transmil}and MHIM-MIL~\cite{tang2023multiple}. We compared our model with ResNet50~\cite{he2016deep}, which was pretrained on ImageNet and later structurally adjusted according to~\cite{lu2021data}, and the original BiomedCLIP~\cite{zhang2023biomed}, which utilized its image encoder directly. Our model, referred to as BiomedCLIP+Adapter, underwent PEFT based on the original BiomedCLIP. The experiments are reported in Table~\ref{table:backbone}.

From the experimental findings, the following conclusions can be drawn: Firstly, within the same MIL methods, our model exhibits markedly superior performance in terms of AUC, sensitivity, and specificity, outperforming the other two image encoders. Secondly, the integration of features extracted through diverse MIL methods consistently yields impressive outcomes, with AUCs exceeding 98.8\%. This quantitative evidence strongly substantiates the notion that the image encoder trained by our framework exhibits significantly enhanced representational power.

Fig.~\ref{fig:roc} presents the Receiver Operating Characteristic (ROC) curves for the binary classification of WSIs, where the features of different image encoders are aggregated using AB-MIL~\cite{ilse2018attention}. The graph distinctly illustrates that our method, which employs features extracted with the trained adapter, maintains a higher true positive rate across a spectrum of false positive rates. This visual representation underscores the fact that the features extracted by our model encapsulate more crucial information, thereby enhancing the overall classification performance.

\textbf{Comparison with MIL methods with frozen pretrained models on FNAC.}
To ascertain the applicability of our method across diverse datasets, we conducted experiments on the publicly available FNAC 2019~\cite{saikia2019comparative} dataset. Table~\ref{table:fnac} presents a comparative analysis of our trained image encoder against ResNet50 and BiomedCLIP when employed with MHIM-TransMIL~\cite{tang2023multiple} on this dataset. Both BiomedCLIP and BiomedCLIP-SimCLR achieved perfect accuracy, which primarily relies on the powerful representation capabilities of the underlying foundation model. This achievement is also indicative of the simplicity of the task, potentially due to the relatively small size of the dataset. Despite achieving 100\% accuracy on the test set, BIomedCLIP-SimCLR demonstrates faster convergence, converging in 93 epochs as opposed to BiomedCLIP’s 106 epochs. This suggests that the proposed adaptation module enhances the representation of task-specific images, and it underscores the versatility of our approach, enabling its application to cytopathology image screening beyond cervical images.

\begin{table}
\centering
\caption{Results with different feature extractors on FNAC dataset.}
\label{table:fnac}
\scalebox{0.9}{
\begin{tblr}{
  cells = {c},
  vline{2} = {-}{},
  hline{1-2,5} = {-}{},
}
Feature extractor & AUC   & Acc   & F1    \\
ResNet50          & 99.78 & 97.67 & 97.96 \\
BiomedCLIP        & 100.00   & 100.00   & 100.00   \\
BiomedCLIP+Adapter & 100.00   & 100.00   & 100.00   
\end{tblr}
}
\end{table}

\subsection{Ablation Experiments}

\begin{table}
\centering
\caption{Ablation results with different filter strategies for contrastive learning.}
\label{table:strategy}
\scalebox{0.9}{
\begin{tblr}{
  cells = {c},
  vline{3} = {-}{},
  hline{1-2,6} = {-}{},
}
Filter    & Con & AUC   & Acc   & F1    \\
          &     & 98.00 & 93.00 & 91.95 \\
          & \checkmark   & 99.02 & 95.67 & 94.82 \\
\checkmark (random) & \checkmark   & 98.26 & 95.17 & 94.21 \\
\checkmark (ours)   & \checkmark   & \textbf{99.18} & \textbf{96.18} & \textbf{95.45} 
\end{tblr}
}
\end{table}

\begin{figure}[t]
    \centering
    \includegraphics[width=7.5cm]{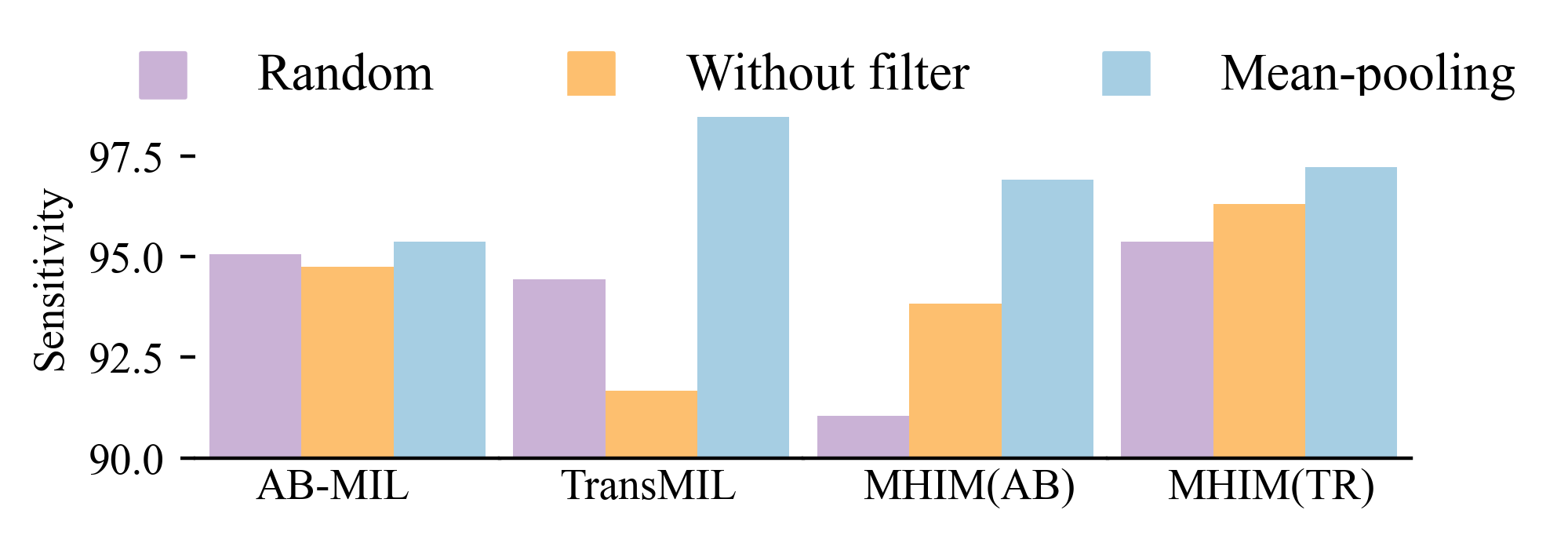}
    \includegraphics[width=7.5cm]{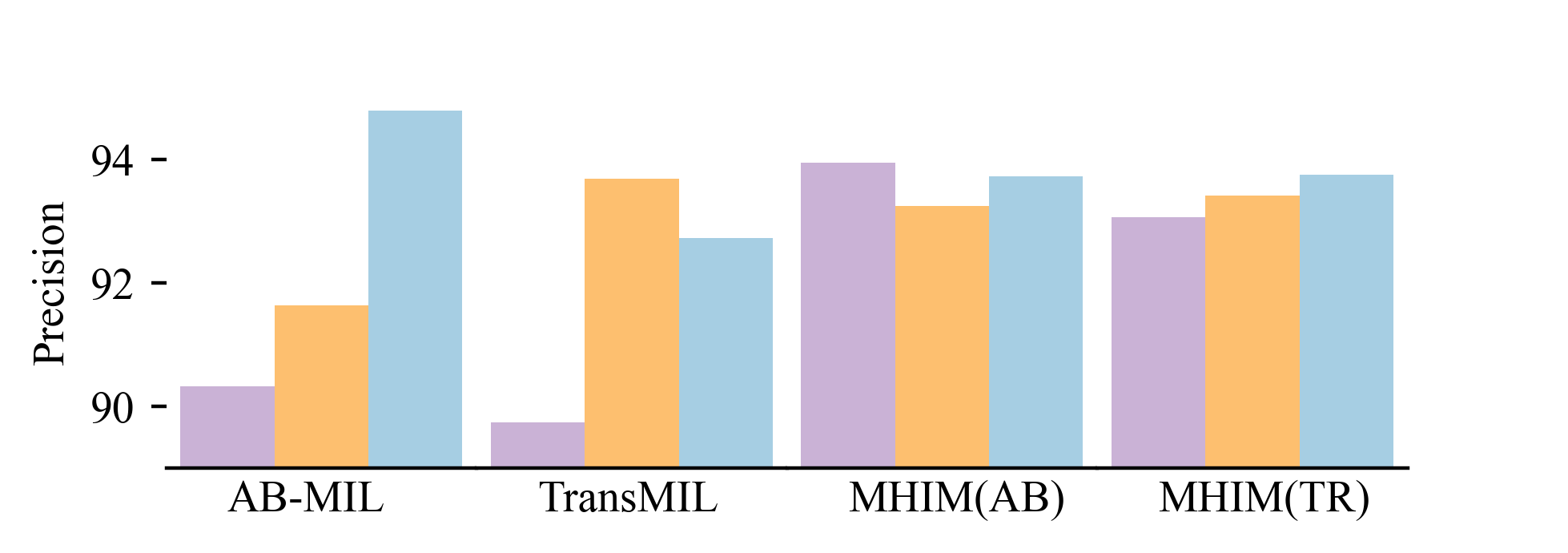}
    \caption{Performance of features extracted by different filter strategies. The extracted features are aggregated by four different MIL methods.}
    \label{fig:filter}
\end{figure}

\begin{figure}[t]
    \centering
        \includegraphics[width=7.5cm]{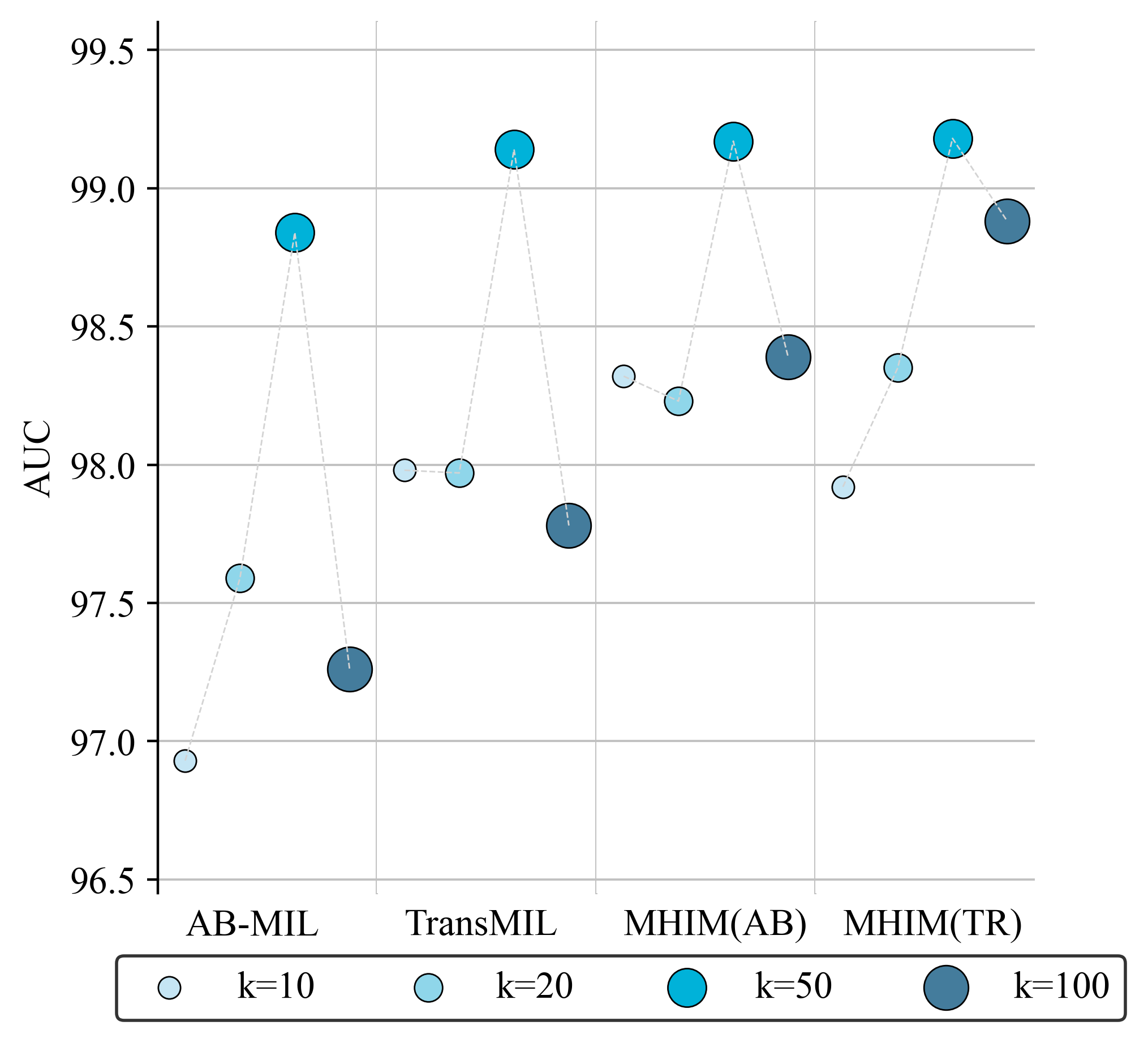}
    \caption{Influence of different settings of $k$.}
    \label{fig:ablation_k}
\end{figure}

We run a number of ablations on the CSD dataset to analyze the proposed framework and discuss it in detail.

\textbf{Filter strategy.}  Given the considerable volume of patches, employing all of them for PEFT would still be impractical due to the excessive time cost. Li~\etal~\cite{li2021dual} previously addressed this issue by adopting a random sampling strategy to reduce the number of patches required for training. However, we argue that this simple strategy may compromise the efficacy of contrastive learning, as the selected patches should be representative of WSIs. Therefore, we propose the MP-based patch filter strategy. Additionally, we have conducted a comparative analysis of the method that utilizes all patches from WSIs.

Table~\ref{table:strategy} shows the aggregation result by MHIM-AB-MIL~\cite{tang2023multiple} with adapters trained by different filter strategies. Experimental outcomes indicate that the features derived from the MP-based strategy surpass those obtained from random sampling and the approach without any filtering. The method without any filter, which trains the adapter with over 1 million images, exhibits inferior performance. This degradation may be attributed to the excessive background information in cytopathology WSIs, which hampers their ability to effectively represent lesion areas.

Fig.~\ref{fig:filter} illustrates the sensitivity and precision of features extracted through different filter strategies under four MIL methods. While the MP-based approach does not achieve the highest precision in the TransMIL and MHIM-AB-MIL methods, the bar chart reveals that our strategy surpasses other methods by improving precision by less than 1\% while enhancing sensitivity by over 5\%. Overall, across various MIL methods, our proposed MP-based filter strategy is more effective in enhancing performance.

\textbf{Filter patch number $k$.} We also investigate the influence of different settings of the filter patch number $k$. Fig.~\ref{fig:ablation_k} reports results and shows that our framework achieves the best performance when $k$ is set to 50.

\textbf{Foundation model.} 
As mentioned before, there are now numerous foundation models available, and in addition to BiomedCLIP~\cite{zhang2023biomed}, we conducted experiments using image encoders from CLIP~\cite{radford2021learning} and PLIP~\cite{huang2023visual}. 
Table~\ref{table:foundation} shows that under our framework, CLIP~\cite{radford2021learning} and PLIP~\cite{huang2023visual} demonstrated exceptional performance, comparable to BiomedCLIP~\cite{zhang2023biomed}. The possible reason for this may be that CLIP was pretrained on a larger dataset and PLIP was trained on a dataset that is more relevant to cervical WSI. Notably, When using MHIM-TransMIL~\cite{tang2023multiple}, PLIP exhibited the best performance, with AUC of 99.69\% and specificity of 98.27\%. In contrast, when employing other MIL methods, CLIP demonstrated superior performance. Overall, regardless of the foundation model used, our framework improves the performance of MIL methods, as compared to the traditional ResNet50~\cite{he2016deep} patch-level feature extractor.

\begin{table}
\centering
\caption{Ablation results with different foundation models.}
\label{table:foundation}
\scalebox{0.9}{
\begin{tblr}{
  cells = {c},
  cell{2}{1} = {r=4}{},
  cell{6}{1} = {r=4}{},
  cell{10}{1} = {r=4}{},
  vline{3} = {-}{},
  hline{1-2,6,10,14} = {-}{},
}
Feature extractor             & MIL      & AUC            & Sens           & Spec           \\
{CLIP\\+Adapter}          & AB-MIL~\cite{ilse2018attention}   & 99.03          & 98.15          & 95.89          \\
                          & TransMIL~\cite{shao2021transmil}& 99.43          & 98.15          & 96.10           \\
                          & MHIM(AB)~\cite{tang2023multiple}& 99.58          & 97.53          & 98.05          \\
                          & MHIM(TR)~\cite{tang2023multiple}& 99.61          & 96.91          & 98.05          \\
{PLIP\\+Adapter}          & AB-MIL~\cite{ilse2018attention}  & 98.89          & 93.21          & 96.54          \\
                          & TransMIL~\cite{shao2021transmil}& 99.33          & 96.91          & 96.11          \\
                          & MHIM(AB)~\cite{tang2023multiple}& 99.45          & \textbf{98.46} & 96.97          \\
                          & MHIM(TR)~\cite{tang2023multiple}& \textbf{99.69} & 96.30           & \textbf{98.27} \\
{BiomedCLIP   \\+Adapter} & AB-MIL~\cite{ilse2018attention}  & 98.84          & 95.37          & 96.32          \\
                          & TransMIL~\cite{shao2021transmil}& 99.14          & \textbf{98.46} & 94.59          \\
                          & MHIM(AB)~\cite{tang2023multiple}& 99.17          & 96.91          & 95.45          \\
                          & MHIM(TR)~\cite{tang2023multiple}& 99.18          & 97.22          & 95.45          
\end{tblr}
}
\end{table}

\section{Conclusion}
\label{sec:conclusion}

In this paper, we propose an efficient detection-free framework for cervical cytopathology WSI classification. In the first stage, we employ an MP-based strategy to filter patches. In the second stage, we use high-risk patches to adapt the frozen image encoder to cervical images through contrastive learning. In the third stage, we extract features using the trained feature extractor and feed them into embedding-based MIL approaches for screening. Experimental results indicate that our proposed method outperforms task-specific approaches for cervical WSIs and MIL methods with a frozen image encoder, achieving SOTA performance on the CSD dataset. The framework can be directly extended to the classification of histopathology WSIs. 
Furthermore, the intricate procedural complexity and the necessity for enhanced interpretability of the methods represent areas in need of improvement.

\section*{Acknowledgements}
This manuscript was supported in part by the National Key Research and Development Program of China under Grant 2021YFF1201202,  the Key Research and Development Program of Hunan Province under Grant 2023SK2029, and the Natural Science Foundation of Hunan Province under Grant 2024JJ5444 and 2023JJ30699. 
The authors wish to acknowledge High Performance Computing Center of Central South University for computational resources.

\bibliographystyle{elsarticle-num} 
\bibliography{refs}






\end{document}